\documentclass[letterpaper, 10pt, conference]{ieeeconf}  

\IEEEoverridecommandlockouts                              

\usepackage{graphicx}
\usepackage{amsmath,amssymb,amsfonts}
\usepackage{algpseudocode,algorithm}
\usepackage{subcaption}
\usepackage{makecell}
\usepackage{cite}
\usepackage{url}
\usepackage{multirow}

\title{\LARGE \bf
Skin-Machine Interface with Multimodal Contact Motion Classifier
}

\author{Alberto Confente$^{1}$, Takanori Jin$^{2}$, Taisuke Kobayashi$^{2}$, Julio Rogelio Guadarrama-Olvera$^{3}$, and Gordon Cheng$^{3}$
\thanks{*This work was supported by JST, CRONOS, Japan Grant Number JPMJCS24K6.}
\thanks{$^{1}$A. Confente is with the Department of Informatics, Technical University of Munich, Germany and with the National Institute of Informatics (NII), Japan
        {\tt\small ge87cuf@mytum.de}}%
\thanks{$^{2}$T. Jin and T. Kobayashi are with the National Institute of Informatics (NII) and with The Graduate University for Advanced Studies (SOKENDAI),
        2-1-2 Hitotsubashi, Chiyoda-ku, Tokyo, 101-8430, Japan
        {\tt\small \{jin, kobayashi\}@nii.ac.jp}}%
\thanks{$^{3}$J. Rogelio and G. Cheng are with the Institute for Cognitive Systems (ICS), Technical University of Munich, Germany
        {\tt\small \{rogelio.guadarrama, gordon\}@tum.de}}%
}

\begin{document}

\maketitle
\thispagestyle{empty}
\pagestyle{empty}

\begin{abstract}

This paper proposes a novel framework for utilizing skin sensors as a new operation interface of complex robots.
The skin sensors employed in this study possess the capability to quantify multimodal tactile information at multiple contact points.
The time-series data generated from these sensors is anticipated to facilitate the classification of diverse contact motions exhibited by an operator.
By mapping the classification results with robot motion primitives, a diverse range of robot motions can be generated by altering the manner in which the skin sensors are interacted with.
In this paper, we focus on a learning-based contact motion classifier employing recurrent neural networks.
This classifier is a pivotal factor in the success of this framework.
Furthermore, we elucidate the requisite conditions for software-hardware designs.
Firstly, multimodal sensing and its comprehensive encoding significantly contribute to the enhancement of classification accuracy and learning stability.
Utilizing all modalities simultaneously as inputs to the classifier proves to be an effective approach.
Secondly, it is essential to mount the skin sensors on a flexible and compliant support to enable the activation of three-axis accelerometers.
These accelerometers are capable of measuring horizontal tactile information, thereby enhancing the correlation with other modalities.
Furthermore, they serve to absorb the noises generated by the robot’s movements during deployment.
Through these discoveries, the accuracy of the developed classifier surpassed 95~\%, enabling the dual-arm mobile manipulator to execute a diverse range of tasks via the Skin-Machine Interface.

\end{abstract}

\section{Introduction}

In recent years, the expectations for robots have shifted from routine tasks in closed environments to general-purpose tasks in more open environments.
This expectation stems from the rapid development of machine learning technology, and in particular, robotic foundation models based on imitation learning technology have been found to robustly execute a wide variety of tasks by learning from demonstrations~\cite{kawaharazuka2024real,black2024pi_0}.
To improve the quality of such models, numerous data samples are be desired based on the belief of the scaling law.

Interfaces for operating robots are essential in collecting data of the performed tasks~\cite{kent2017comparison,darvish2023teleoperation}.
In other words, an operator needs to be familiar with the intent and procedure of the target tasks and can perform them through the robot's body, executing action patterns that the robot can perform.
There are two major directions in the development of such an interface to operate the robot's body:
\textit{i) ``direct'' operation with injective mapping from the operator's whole body}; and
\textit{ii) ``abstract'' operation with surjective mapping from the operator's limited body}.

\begin{figure}[tb]
    \centering
    \includegraphics[keepaspectratio=true,width=0.96\linewidth]{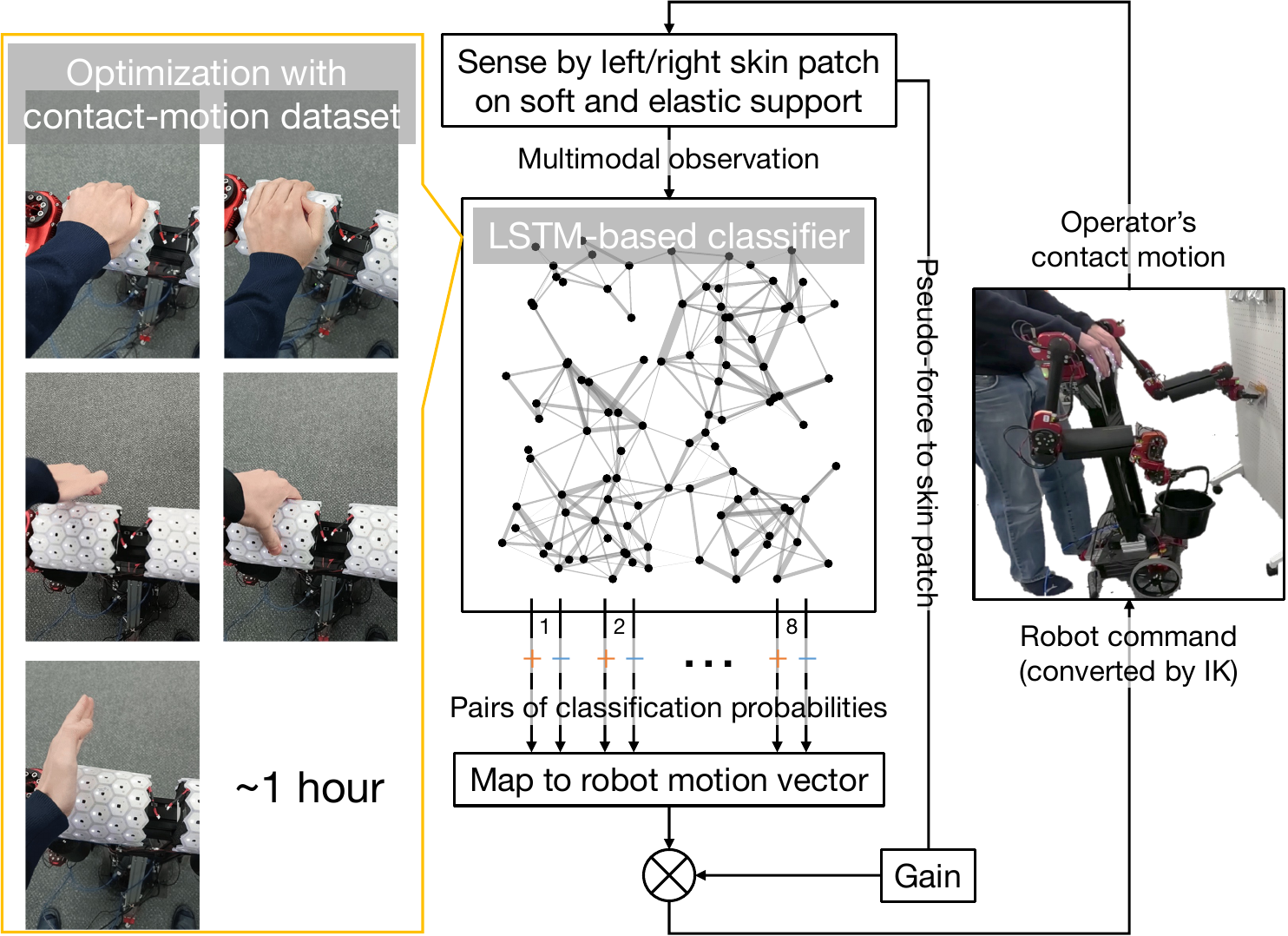}
    \caption{Proposed Skin-Machine Interface (SMI) framework}
    \label{fig:framework}
\end{figure}

To obtain high-quality demonstrations, \textit{i) direct operation with injective mapping} has become important in recent years.
For a robotic manipulator, if the interface has a body structure analogous to it, it can be operated directly at the joint-space level with high accuracy~\cite{ito2022efficient,fu2024mobile}.
For a humanoid robot, human skeletal data, which can be measured by motion capture systems and/or camera images, is retargeted to match its body structure, making the operator manipulate it without physical resistance and/or geometrical constraints of the interface~\cite{he2024learning,bertrand2024high}.
On the other hand, due to the characteristics of injective mapping, the degree of freedom available for operation is limited.
For example, it is inconvenient to operate the base of a mobile manipulator by introducing additional interfaces~\cite{ito2022efficient}, or forcibly pushing/pulling with the torso~\cite{fu2024mobile}.
This is also vulnerable to sensing failures (e.g., observation loss by occlusion) due to no bypass.

Another direction, \textit{ii) abstract operation with surjective mapping}, is expected to achieve a high degree of freedom of robot motions while using only a small part of the operator's body for operation.
The most obvious example should be the brain-machine interface, which enables arbitrary robot manipulation by measuring high-dimensional information from a very limited operating area, i.e., the brain activity, and by extracting diverse operation intentions latent in the information~\cite{tonin2021noninvasive,ferrero2024brain}.
The challenge is to measure high-dimensional information and extract operation intentions in operating more parts of the robot from a limited number of parts of the operator.
Nevertheless, the interface can be highly versatile without being structurally constrained as in the first direction.
In addition, as a side effect of restricting the operating area, it is expected that people with any kind of abilities can operate the robot.

In the context of teaching physical movements to individuals, the instructor typically guides the learner through the execution of the desired motions by providing tactile feedback.
In essence, the sense of touch possesses remarkable information resolution and can serve as an interface with which the operator is accustomed.
From this point of view, previous studies have used artificial skin sensors as a kind of interface to trigger simple contact motions (e.g., tapping) to switch robot state machines~\cite{wong2022touch}; to perform direct teaching by contact force when combined with impedance control~\cite{cheng2019comprehensive}; or both~\cite{jung2024touch}.
Note, however, that the latter involves data collection under dynamics different from those of robot automation, and the accumulation of errors may cause the learned model to fail the target tasks.

With the abovementioned, our study aims to develop a general-purpose Skin-Machine Interface (SMI) as the abstract operation.
That is, our SMI classifies complex contact motions made with small dual-hand movements and links them to robot operations, as illustrated in Fig.~\ref{fig:framework}.
Since it is noninvasive and less noisy compared to measuring brain activities, and the information measured by skin sensors should be close to the operator's intentions, we can expect the relative ease of extracting the information for operations through this interface.
Although the intuitiveness of the interface is inferior to that of the direct operation, we emphasize the importance of the high degree of freedom for operating highly redundant robots.

The most important factor in developing this interface is the accurate classification of complex contact motions of the human operator.
Recent advances in deep learning have made it possible to train accurate classifiers in a data-driven manner~\cite{mohammadi2024deep}, but this often requires large models and datasets.
For the usage in the interface targeted in this study, large models are not appropriate from the viewpoint of computational costs, and large datasets are also impractical for casually registering a wide variety of contact motions.

In order to achieve high classification accuracy under such constraints, it is important to design appropriate hardware and software.
In this paper, we experimentally reveal the following two necessary requirements.
The first is a comprehensive encoding of multimodal sensing information about tactile sensation, that is, the correlation among modalities can complement missing information and yield noise robustness.
The other is the softness (and elasticity) of the supports on which the sensors are mounted, which activates the measurement of acceleration information, one of the modalities, and strengthens the correlation among modalities.
We show that these innovations enable classification of 17 types of contact motions with an accuracy of more than 95~\%, only with a lightweight model (less than one million parameters) and a small-sized dataset (about one hour time-series length).
We also demonstrate that a dual-arm mobile manipulator can perform various tasks by appropriately translating the classification results into robot control commands.

\section{Proposed SMI framework}

\subsection{Target hardware}

\begin{figure}[tb]
    \centering
    \includegraphics[keepaspectratio=true,width=0.84\linewidth]{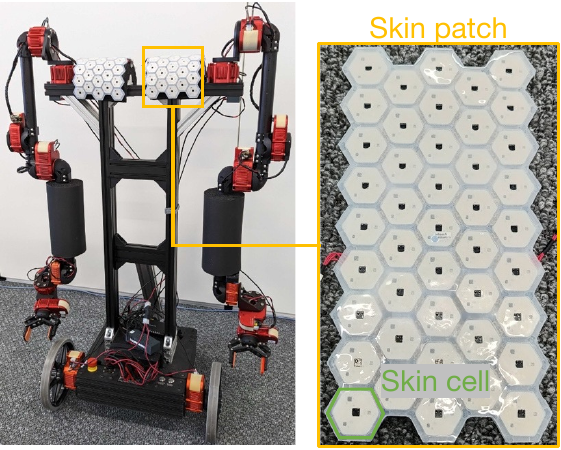}
    \caption{Dual-arm mobile manipulator with skin sensors}
    \label{fig:robot}
\end{figure}

Although the framework proposed in this study (see Fig.~\ref{fig:framework}) is generalizable, we introduce each module by referring to the actual implementation on the robot used in this paper for the sake of ease of understanding.
The robot is a dual-arm mobile manipulator, as shown on the left of Fig.~\ref{fig:robot}.
It consists of two 6DOF arms with a 1DOF two-finger gripper for each, and a two-wheel differential drive mobile base.
These are driven by modular actuators manufactured by Hebi Robotics.
The torso has no DOFs, but its shoulders are equipped with two skin sensor patches (developed by inTouch Robotics) on both sides.
Each patch contains 43 cells, each of which measures the sense of touch (see the right of Fig.~\ref{fig:robot}), and is installed on a soft and elastic elastomer support (more specifically, Aeroflex is employed).

Since this robot system is left-right symmetrical, for simplicity, the framework is constructed so that each patch operates on one side.
In other words, to operate the system, eight dimensions should be handled discriminatively from a single interface:
the displacement of the end-effector's 6D pose; the opening/closing amount of the gripper; and the rotational speed of the wheel.
Note that the 6D pose is controlled by a quadratic programming-based inverse kinematics.
The recent interface that has the same (or analogous) shape as the target manipulator often suffers from operating the gripper or wheel.
In order to compensate for this lack of operability, supplementary interface(s) might be needed~\cite{ito2022efficient}, or the base should be made backdrivable so that the operator can move it directly (although the true action data cannot be obtained in that case)~\cite{fu2024mobile}.
In contrast, the proposed SMI has the capability to operate all of them sufficiently by actively utilizing its spatiotemporal multimodal information.
To this end, the system configurations, a learning algorithm, and the way of collecting data are described in the following sections.

\subsection{Multimodal skin sensor}

Each of the $K=43$ cells in the left/right skin sensor patch contains three types of sensors:
three force sensors that measure the vertical forces at three points on the cell; a proximity sensor that measures the distance to an object placed in the center, and a three-axis acceleration sensor.
Note that the value of the proximity sensor increases as the object gets closer so that it can be easily converted into a pseudo-force~\cite{cheng2019comprehensive}.
These sensor values are low-pass filtered and normalized at the firmware level in the range $[0, 1]$ and published in 100~Hz from its ROS node.
The data measured from each patch at time step $t$ can be summarized as follows:
\begin{align}
    o_t &= [o_{t,1}^\top, o_{t,2}^\top, \ldots, o_{t,K}^\top]^\top
    \\
    o_{t,i} &= [f_{t,i,z1}, f_{t,i,z2}, f_{t,i,z3}, p_{t,i,z}, a_{t,i,x}, a_{t,i,y}, a_{t,i,z}]^\top
\end{align}
where, $f_*$ denotes the force sensor, $p_*$ the proximity sensor, and $a_*$ the acceleration sensor values, where $x,y,z$ are the components for each cell coordinate system.
$o_t \in \mathcal{O} = [0, 1]^{7 \times K = 301}$ holds.
For simplicity, this paper does not use the geometric arrangement of cells, but rather aggregates them into a simple 301-dimensional vector.
In order to construct a common dataset for the left and right patches, however, the indices of cells in the two patches are aligned to be mirrored geometrically.

A pseudo-wrench applied to the patch can be calculated by integrating the force sensation of each cell, considering its geometric arrangement, as was done in previous studies~\cite{cheng2019comprehensive}.
In this paper, the translational component of the wrench, i.e., the applied force, is utilized later, as defined below.
\begin{align}
    F_t = \frac{k_f}{3} \sum_{i=1}^{K} R_i (f_{t,i,z1} + f_{t,i,z2} + f_{t,i,z3})
    \label{eq:F}
\end{align}
where, $R_i$ is a rotation matrix to convert the coordinate system of $i$-th cell to the patch origin one, and $k_f$ is a calibrated scale to map to the true force value.
Note that the proximity sensor value $p_{t,i,z}$ is also used to calculate the pseudo-force, but in this study, it is excluded because its change is small and the force sensor values are dominant.

\subsection{Learning of classifier}

The sequences of $o_t$, measured from the left/right patch, should represent the operator's contact motions.
We train a model that classifies them in real-time.
Specifically, given a dataset $D$ containing $N$ trajectories $\tau_n = [o_{n,1}, o_{n,2}, \ldots, o_{n,T_n}]$ in pairs with the target contact motion labels $c_n$ ($n=1, 2, \ldots, T_n$), we solve the following minimization problem with negative log-likelihood of a trainable classifier model $p_c$ with $\theta$ the trainable parameters.
\begin{align}
    \theta^\ast = \mathop{\mathrm{arg\,min}}_\theta \mathbb{E}_{(\tau_n, c_n) \sim D}[- \sum_{t=1}^{T_n} \ln p_c(c_n \mid o_{n,\leq t}; \theta)]
\end{align}
where, $o_{n,\leq t} = [o_{1}, \ldots, o_{t}]$ is the segment of $\tau_n$ from the initial to $t$-th steps.
The details of $D$ is described later.

Recent time-series processing models with deep learning require approximating $o_{\leq t}$ as the corresponding feature $h \in \mathcal{H} = \mathbb{R}^{|\mathcal{H}|}$.
This paper employs a long short-term memory (LSTM)~\cite{hochreiter1997long}, which is categorized into recurrent neural networks, another representative model that enables fast inference.
Using LSTM $f$ with $\phi$ the trainable parameters, $h$ is updated as follows:
\begin{align}
    o_{\leq t} \simeq h_t = f(h_{t-1}, o_t; \phi)
\end{align}
where, it is common practice to initialize $h_0 = 0$, followed in this paper.
By feeding $h_t$ to $p_c$ instead of $o_{\leq t}$, $p_c$ can capture and classify time-series changes in the contact motions.

Since such a structure with LSTM is often overlearned, this paper introduces an auxiliary objective to regularize it~\cite{gangwani2020learning}.
That is, a predictor model, $p_p$ with $\eta$ the trainable parameters, are additionally trained to stochastically predict $o_{t+1}$ from $h_t$.
In this way, $\mathcal{H}$ is corrected so that the features of the time-series data are appropriately embedded, and to some extent, it is expected to mitigate the collapse of $\mathcal{H}$ to excessively focus on the classification of the given dataset, which basically loses generalization performance.

In summary, the target optimization problem is finally given below.
\begin{align}
    \theta^\ast, \phi^\ast, \eta^\ast &= \mathop{\mathrm{arg\,min}}_{\theta, \phi, \eta}
    \mathbb{E}_{(\tau_n, c_n) \sim D} \Biggl[
    - \ln p_c(c_n \mid h_{T_n}; \theta)
    \nonumber \\
    &- \sum_{t=1}^{T_n-1} \{\ln p_c(c_n \mid h_t; \theta)
    \!+\! \gamma \ln p_p(o_{n,t+1} \mid h_t; \eta)\} \Biggr]
    \\
    \mathrm{s.t.} \
    h_t &= f(h_{t-1}, o_t; \phi) \ (t=1, 2, \ldots, T_n)
    \nonumber
\end{align}
where, $\gamma \geq 0$ is the gain of the regularization by the predictor model ($\gamma=0.02$ in this paper).
Note that this regularization experimentally stabilized the learning by reducing the effects of random seed, although the final classification accuracy did not change.

\begin{figure}[tb]
    \centering
    \includegraphics[keepaspectratio=true,width=0.96\linewidth]{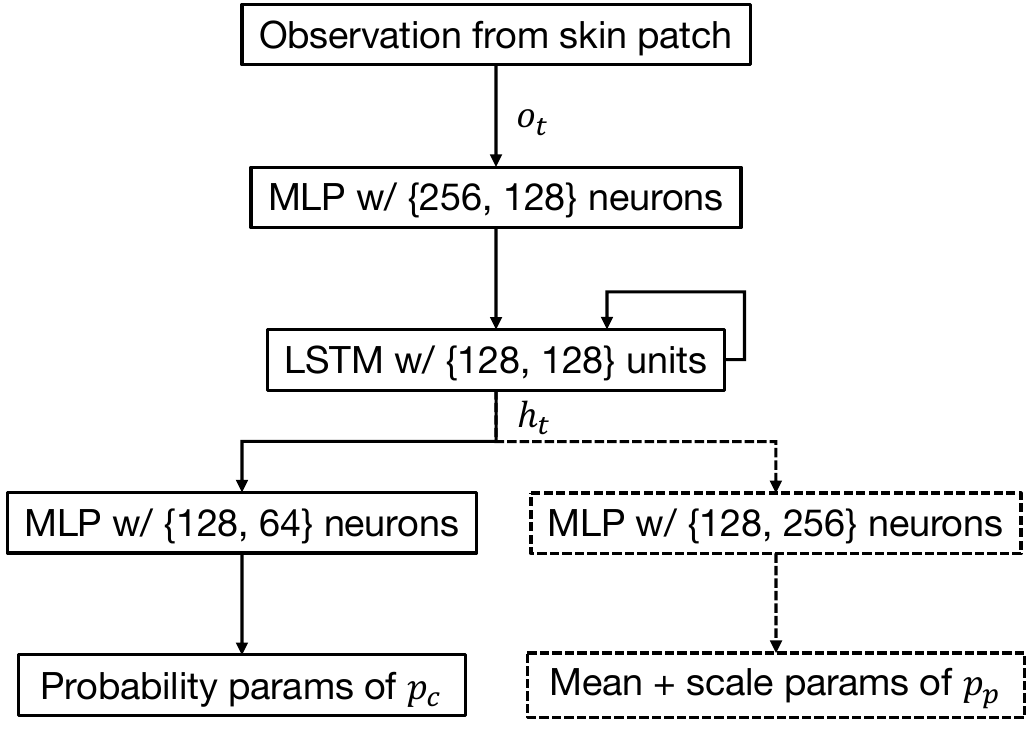}
    \caption{Model structure (dashed parts are only for training)}
    \label{fig:model}
\end{figure}

The actual implementation used is shown in Fig.~\ref{fig:model}:
$o_t$ is encoded into the feature by a multi-layer perceptron (MLP) module before passing it to LSTM;
$h_t$ outputted from the LSTM is formatted by separate MLP modules before passing it to the classifier and predictor models;
and the probability parameter in the classifier model and the mean and scale parameters in the predictor model are finally obtained, respectively.
The number of parameters in this classifier (including the predictor) is 681,523, and its computation time is adequately less than five milliseconds.

\subsection{Data collection}
\label{subsec:data}

Before constructing the dataset for training the classifier as described above, we first consider the contact motions to be registered.
Considering that the probability $[0, 1]$ of the classifier is used to generate eight-dimensional motions with each patch, it is not possible to move arbitrarily in both directions without preparing contact motions for the forward and reverse directions, respectively.
In other words, $8 \times 2 = 16$ different classes must be registered.
In addition to this, a class for stopping the robot's motion is also necessary to prevent accidental operation.

In total, 17 different classes are registered, and their selection requirements are the following.
\begin{enumerate}
    \item The contact motions are intuitively and conceptually linked to the robot movements.
    \item The contact motions are easily distinguishable to simplify training and to enhance classification accuracy.
    \item It is preferred to select a combination of classes that minimizes the hand movement of the operator to avoid unnecessary fatigue.
\end{enumerate}
Although it is impossible to uniquely specify the classes only by these, this paper empirically registers the contact motions in Table~\ref{tab:mapping}.
In the classes named \textit{Torque}, the whole surface of the hand is placed on the sensor with all fingers in parallel, and torque is applied in the direction described in the table.
On the other hand, in the classes named \textit{Grab}, while the palm of the hand is not in contact with the sensor, the thumb and the other four fingers grasp the sensor, and the hand tries to move in the direction described in the table.

\begin{table}
    \caption{Mapping from contact to robot motions}
    \label{tab:mapping}
    \centering
    \begin{tabular}{ll}
        \hline\hline
        \textbf{Contact motion classes} & \textbf{Associated robot motion commands}
        \\
        \hline
        Torque left / right & Linear velocity on the x-axis
        \\
        Torque forward / backward & Linear velocity on the y-axis
        \\
        Torque clock / anticlock & Linear velocity on the z-axis
        \\
        \hline
        Grab left / right & Angular velocity on the x-axis
        \\
        Grab forward / backward & Angular velocity on the y-axis
        \\
        Grab clock / anticlock & Angular velocity on the z-axis
        \\
        \hline
        Touch outside / inside & Gripper opening/closing
        \\
        Push / Pull & Wheel rotation
        \\
        \hline
        No touch & No movement
        \\
        \hline\hline
    \end{tabular}
\end{table}

The dataset is collected to register the 17 classes defined in this way.
Basically, for simplicity of implementation, the sequence length of the $n$-th trajectory is aligned at $T_n=375$ (about four seconds), which can include any contact motion completely.
Each trajectory is assumed to belong to a single class, as implied when introducing the classifier training algorithm.
Note that these limitations might be relaxed by a combination of clustering by sequence length and interpolation by padding, and/or by trajectory segmentation.

Three tricks are incorporated under this setting.
\begin{enumerate}
    \item Measure the contact motions on the left and right patches and add them to a common dataset.
    This allows the classifier to absorb left-right gaps in operator, patch characteristics, and mounting positions, and to correctly classify even if the measured data from either patch is input to the common classifier trained using the common dataset.
    \item Set the initial state of the trajectory to be in the middle of a randomly-selected contact motion, from which the target contact motion is executed.
    The label of this trajectory is that of the target.
    In this way, the classifications become probabilistic near the boundaries of the contact motions, resulting in faster and smoother transitions of the classification results by the classifier.
    \item Collect and train the dataset iteratively.
    In other words, when the trained classifier is deployed on the real robot, classes that are prone to misclassification are identified, and such data is added repeatedly.
    In this way, a satisfactory classification accuracy can be achieved with a minimum amount of data, without collecting a huge amount of data.
\end{enumerate}
By incorporating the above tricks, 1038 trajectories (i.e., about one hour) were eventually collected in total.

\subsection{Map from contact to robot motions}

Based on the above design and the outputs from the classifier model trained on the collected dataset, the robot's motions are generated.
Note again that since each output of the classifier is a class probability, $[0, 1]$, the robot cannot move freely in both forward and reverse directions with only one output.
Then, $i$-th robot's motion, $a_i$, is determined by combining the corresponding pair of two classes, $(c^{i+}, c^{i-})$.
\begin{align}
    a_{i,t} = k_i (p_c(c^{i+} \mid h_t; \theta) - p_c(c^{i-} \mid h_t; \theta))
\end{align}
where, $k_i > 0$ denotes the gain.

Now, we focus on the fact that the classification result alone cannot fully reflect the operator's intent.
Basically, the more forceful the operator's contact is, the more quickly the robot should move, but this information is not included in the classification results (although it should be reflected to some extent, since the weaker the force, the less difference in the contact motions).
Therefore, $k_i$ is designed to be proportional to the operator's force, namely $F_t$ in eq.~\eqref{eq:F}.
\begin{align}
    k_i = \hat{k}_i F_t
    \label{eq:gain}
\end{align}
The constant gain $\hat{k}_i > 0$ is empirically adjusted to optimize operability in each dimension.
In addition, the deadzone and saturation are introduced to improve operability and safety, and their thresholds are also adjusted empirically.
Note that these were determined based on the subjective usability of the interface when the user tries it out, but their optimization would be possible by preparing tests to measure the robot's performance through the interface.

\section{Investigation of requisite conditions for improving the classifier}

\subsection{Learning setup}

The learning setup for the proposed model, as already shown in Fig.~\ref{fig:model}, is summarized here.
This is common to the following comparisons, and experiment-specific conditions are described in each section.

First, 80~\% of the collected dataset is divided into a training dataset and the remaining 20~\% into a validation dataset.
From these datasets, trajectory-label pairs with a batch size of 128 are sampled in a shuffled order.
The trajectories are fed into the model in order from the beginning, and the losses are calculated.
To increase the update frequency, the model parameters are updated by a stochastic gradient descent optimizer after every 100 steps (or at the end of trajectories) with the sum of the losses up to that time.
Note that, after the update, the computational graph for the internal state of LSTM is retained to maintain the long-term impact of the inputs, and it is reset only at the beginning of trajectories.
The optimizer employed in this study is AdaTerm~\cite{ilboudo2023adaterm}, which is robust to gradient noise and outliers, to mitigate the negative effects of complex gradients introduced by LSTM.
AdaTerm uses its default settings, suggested in the paper and a repository available on GitHub, except for the learning rate, which is reduced to $5 \times 10^{-4}$ to stably train LSTM.

The performance of the trained model is evaluated by averaging the classification accuracy (ACC) over the validation dataset.
ACC is calculated as follows:
\begin{align}
    \mathrm{ACC} = \frac{\sum_{n}^{N^\mathrm{valid}}\sum_{t}^{T} \mathbb{I}(c_n = \mathop{\mathrm{arg\,max}}_{c} p_c(c \mid h_{n,t}; \theta))}{N^\mathrm{valid}T}
\end{align}
where, $N^\mathrm{valid} = 0.2 N$ denotes the number of trajectories in the valid dataset, and $\mathbb{I}(b)$ denotes the indicator function, which returns one if $b$ is true; and zero otherwise.
Note that the beginning of the trajectory starts with a different class, as mentioned in Section~\ref{subsec:data}, so it is not realistic to make the ACC of the model reach 100~\%.
It is also remarked that we need to evaluate the statistical performance to absorb randomness depending on the specified random seed, so nine models with different seeds are trained for each comparison.

\subsection{Multimodality}

\begin{figure}[tb]
    \centering
    \includegraphics[keepaspectratio=true,width=0.96\linewidth]{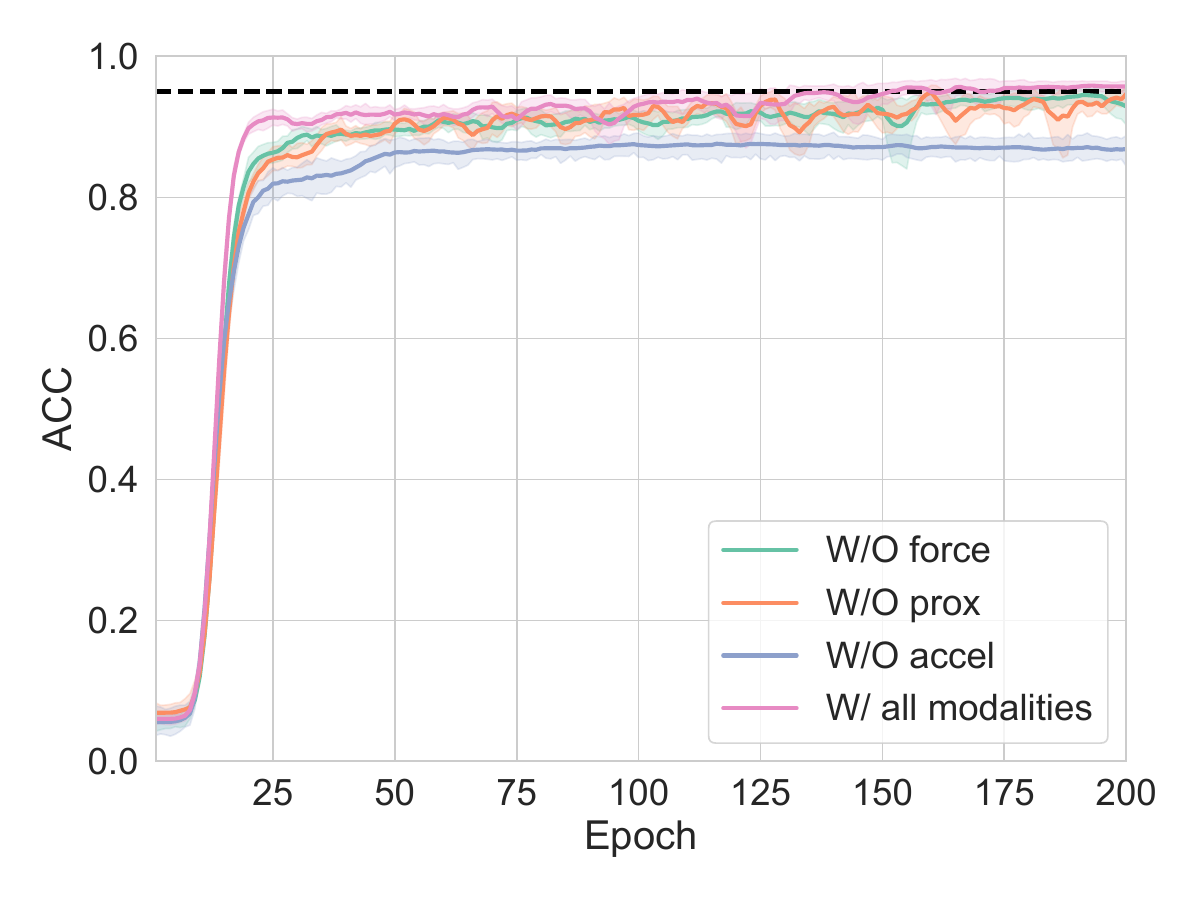}
    \caption{Comparisons of modalities (95~\% ACC denoted by the black dashed line)}
    \label{fig:learn_modality}
\end{figure}

The classification accuracy for all 17 classes is compared while excluding one of the skin sensor modalities.
In this experiment, the training takes 200 epochs to complete.
The learning results are shown in Fig.~\ref{fig:learn_modality}.

The classification accuracies in all cases are similar until around 20 epochs, exceeding 70~\%.
Afterwards, the most characteristic case is when the acceleration sensor is excluded, where the classification accuracy converged to no more than 90~\%.
This is because only the acceleration sensor can measure horizontal tactile information for each cell and directly classify the contact motions acting in that direction.
The vertical tactile information alone can classify motions from its distribution information, and thus, this case did not cause an extreme decrease in classification accuracy.
However, the acceleration sensor is critical to improve the reaction speed by classifying from shorter time-series data.

The next characteristic case is when the proximity sensor is excluded.
The trend is generally the same as when the force sensor is excluded, but the instability of the learning was more pronounced.
This is probably due to the fact that the proximity sensor provides a kind of attention mechanism, which discriminates cells that are touched by the operator and preferentially uses them for classification.
For example, since \textit{Torque} and \textit{Grab} motions in Table~\ref{tab:mapping} are with different touches, the use of proximity sensors would facilitate this distinction and mitigate the negative effects of other cells that are not being touched sufficiently.
Although this distinction can also be made using force sensors, the accuracy may not be sufficient because the force applied is not constant.

Finally, when the force sensor is excluded, the classification accuracy was not so low with less instability than we expected.
This is because the acceleration sensor would complement the information from the force sensor.
However, the accuracy is lower than that of the direct force measurement, and there is a delay, resulting in a significant decrease in classification accuracy.

Compared to these ablation cases, when all modalities are taken into account, the classification accuracy improved earlier and continued to improve steadily, eventually achieving the highest classification accuracy (over 95~\%).
Thus, it can be concluded that in order to stably achieve high classification accuracy in the proposed framework, it is essential to actively utilize all modalities of the skin sensors.

\subsection{Soft and elastic support}

Then, to take full advantage of this multimodality, we consider the requirement on the hardware level.
The above experiment suggests that the acceleration sensor plays an important role.
To activate it, the sensor itself should move in response to the contact motions, and if that movement is too small, it will be buried in the noise of the acceleration sensor.
In addition, it must naturally return to its initial position when no force is applied because, of course, it cannot continue to perform the operation if it remains moving in response to the contact motions.
In this regard, our robot system has the skin sensors mounted on soft and elastic elastomer supports, which meet the requirements for the acceleration sensor to function, as also suggested in the previous work~\cite{al2024digetac}.

\begin{figure}[tb]
    \centering
    \includegraphics[keepaspectratio=true,width=0.96\linewidth]{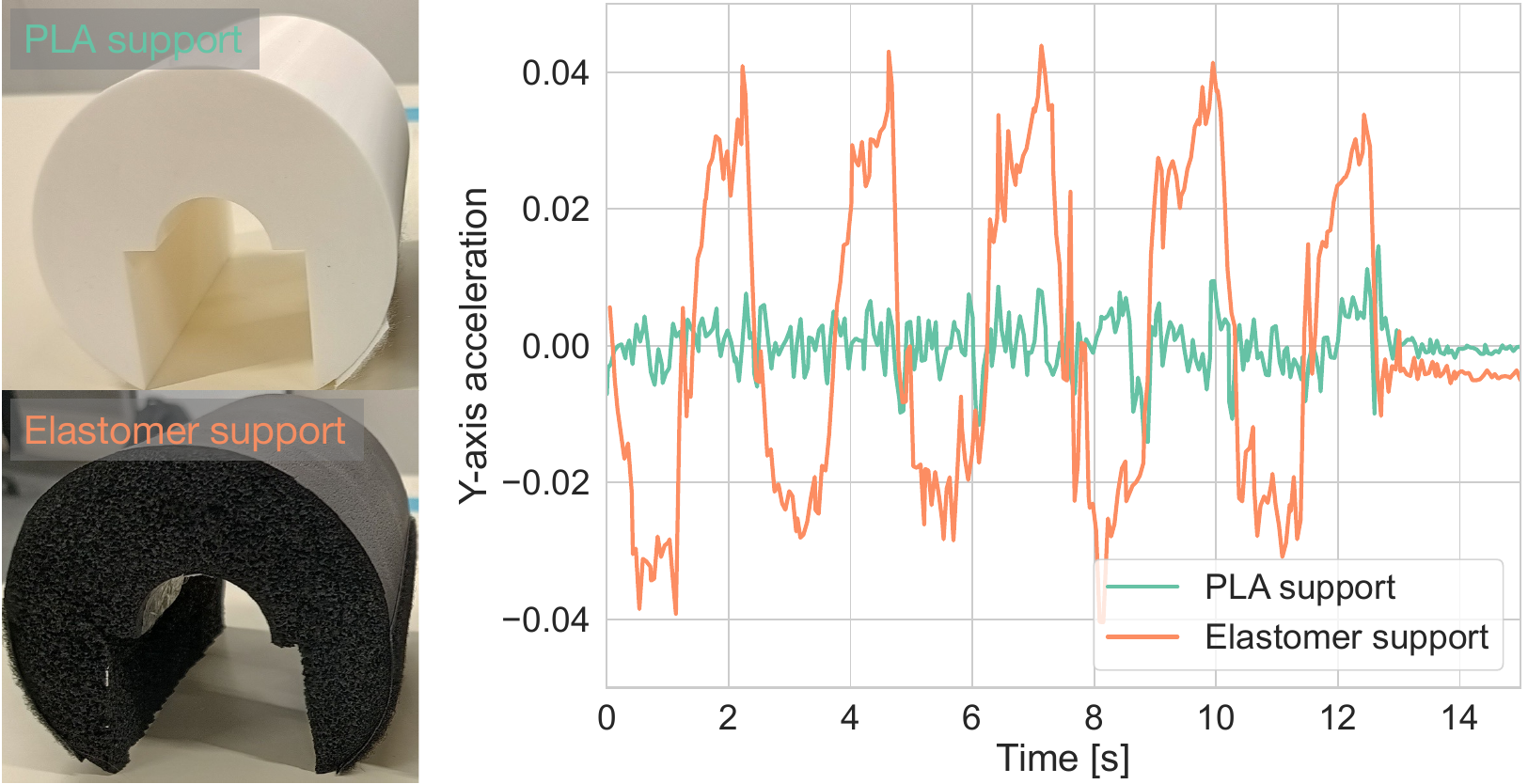}
    \caption{Difference of supports where skin sensors are mounted}
    \label{fig:support}
\end{figure}

To verify the necessity of the soft and elastic support, we printed a rigid PLA support of the same shape using a 3D printer (see the left of Fig.~\ref{fig:support}), and applied $y$-axis forces to the skin sensors placed on the supports.
The result is depicted in the right of Fig.~\ref{fig:support}, showing that the PLA support hardly responds to the force, and noise was dominant.
In contrast, with the elastomer support, the direction of the applied force can be determined from the acceleration sensor value.

Thus, the softness of the support activates the acceleration sensor, creating a strong correlation with the force and allowing it to pick up horizontal tactile information to the cell that cannot be measured directly by other modalities.
The deformation that occurs when gripping the skin patch can also create a gap between the patch and the hand, which is captured by the proximity sensor.
Thus, the soft (and elastic) support is expected to mitigate the influence of noise on each sensor because each modality is strongly correlated with each other.
We additionally examine the effect of these different supports on classification accuracy.
Because different supports produce different data distributions, the respective datasets of the same size were collected separately.
In order to clarify the results of interest, we selected six classes (i.e., the ones for \textit{Torque} motions), where the soft support contributes significantly to the classification, with 25 sequences for each.

\begin{figure}[tb]
    \centering
    \includegraphics[keepaspectratio=true,width=0.96\linewidth]{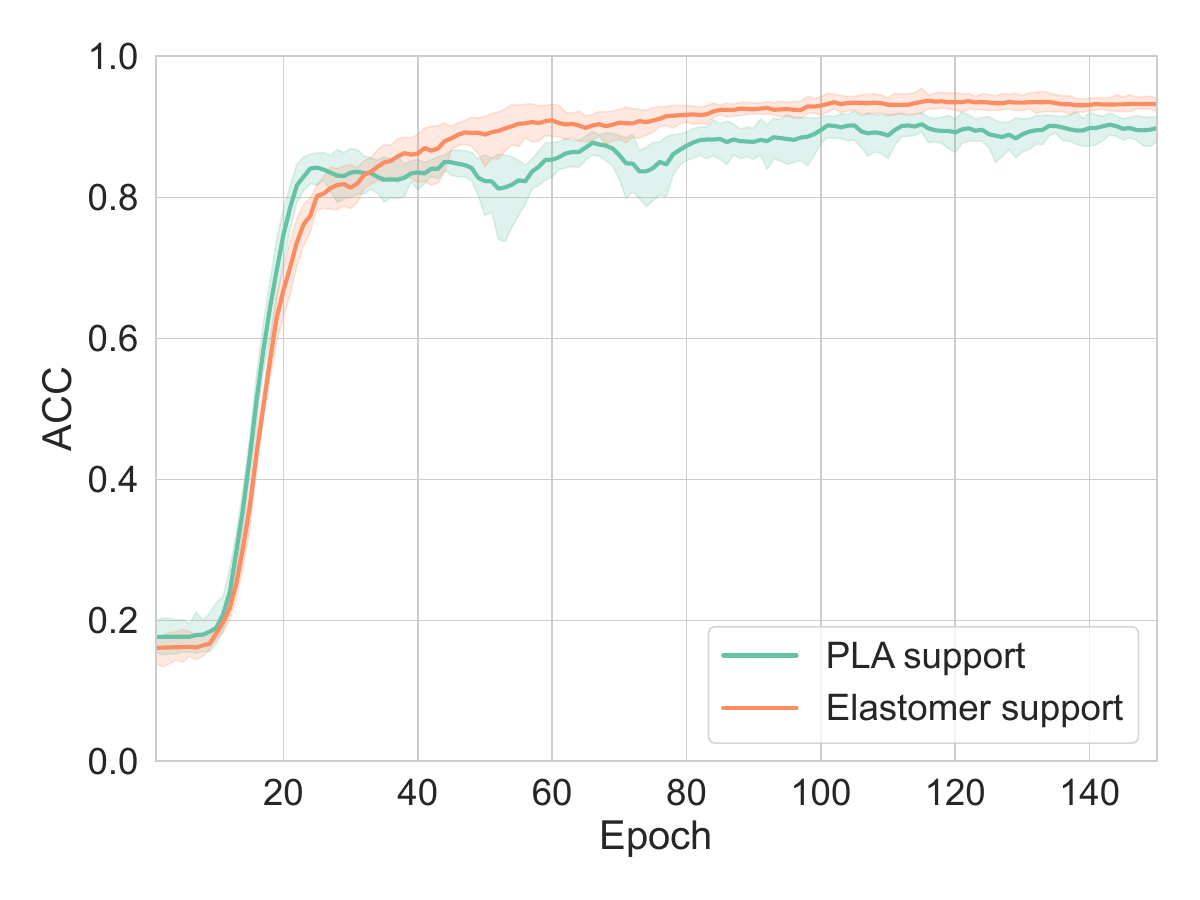}
    \caption{Comparisons of support materials}
    \label{fig:learn_support}
\end{figure}

The training results for 150 epochs are shown in Fig.~\ref{fig:learn_support}.
As expected, classification accuracy was low with the PLA support, and learning became unstable.
On the other hand, by focusing only on the early stage of training, it appears that the classification accuracy with the PLA support improved earlier compared to the case with the elastomer support.
This may be due to the difference in datasets, but it is more likely due to the correlation among multimodal sensors.
In other words, extra learning might be required for capturing the strong correlation between multimodal sensors with the elastomer support, then, the model could utilize it for classifying the contact motions more reliably.

As a remark, we tried to operate the robot with the model for the PLA support, and as expected, the robot could not be operated at all.
One reason is obviously insufficient classification accuracy, but another reason is the more direct interaction between the operator and the robot.
That is, when the operator manipulates the robot, the robot starts to move with acceleration, and this reaction affects the operator through the rigid body.
This effect is different from that of the dataset collection and causes a distribution shift in the operator's contact motions.
In addition, the robot motions also appear as noise to the sensors, as indicated in just before the end of Fig.~\ref{fig:support}, where the robot oscillated when the operator released the robot and that oscillation was confirmed as a large acceleration with the PLA support.
On the other hand, the soft (and elastic) support physically dampens the reactions caused by the robot's movements, thus minimizing the effects of such a distribution shift.
As a result, the proposed framework allows the robot to be operated as desired by the operator, as explained in the next section.

\section{Demonstrations}

\subsection{Results}

\begin{figure*}[tb]
    \centering
    \includegraphics[keepaspectratio=true,width=0.92\linewidth]{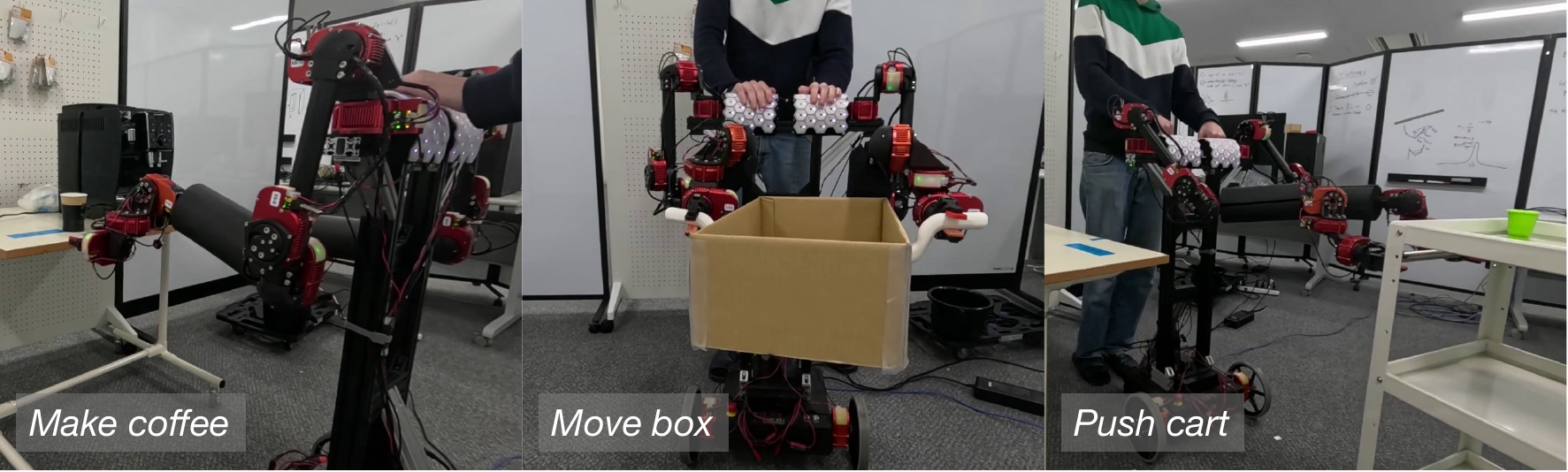}
    \caption{Tasks demonstrated}
    \label{fig:task}
\end{figure*}

\begin{figure*}[tb]
    \begin{subfigure}[b]{0.32\linewidth}
        \centering
        \includegraphics[keepaspectratio=true,width=\linewidth]{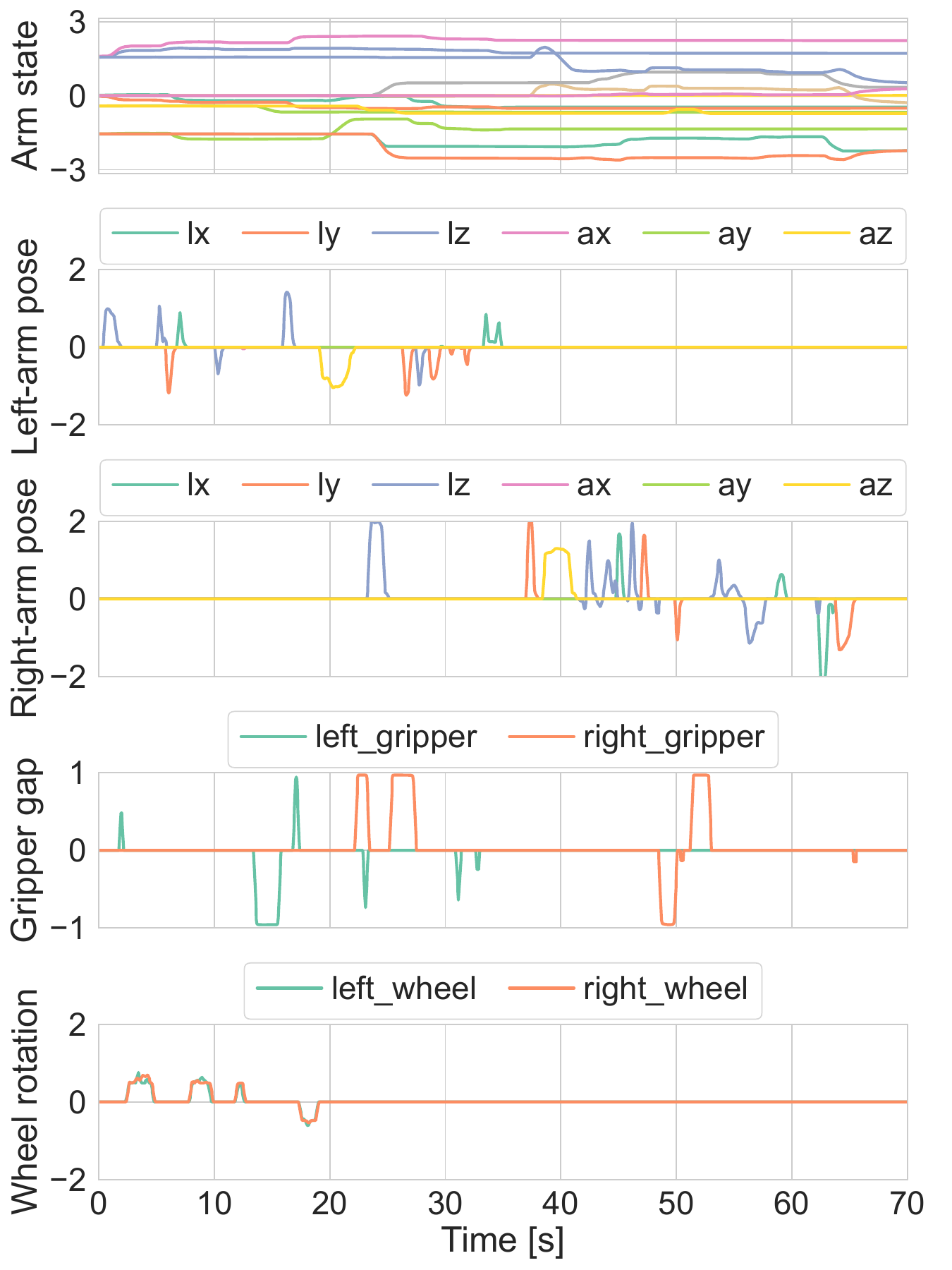}
        \subcaption{\textit{Make coffee}}
        \label{fig:demo_make_coffee}
    \end{subfigure}
    \begin{subfigure}[b]{0.32\linewidth}
        \centering
        \includegraphics[keepaspectratio=true,width=\linewidth]{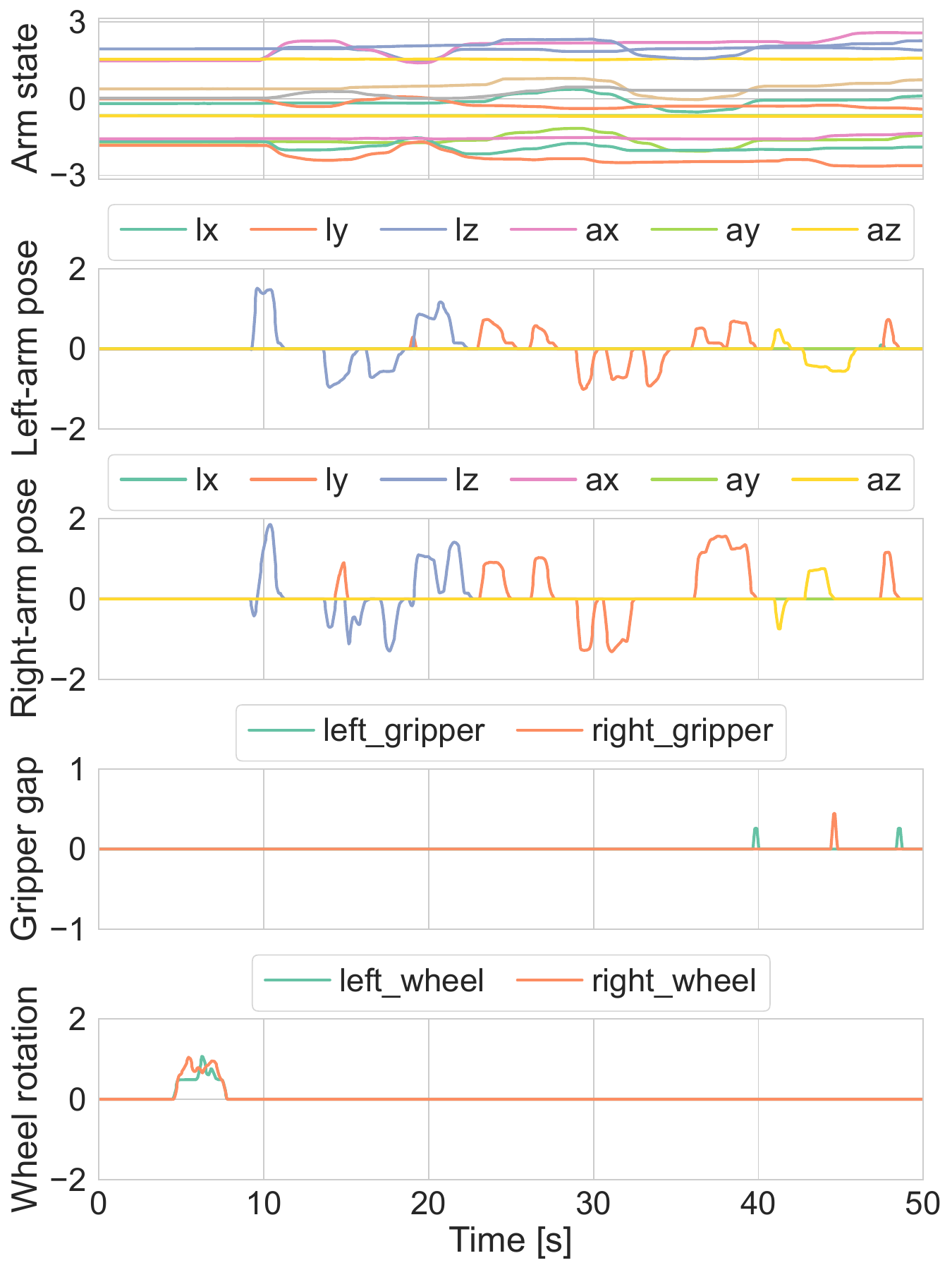}
        \subcaption{\textit{Move box}}
        \label{fig:demo_move_box}
    \end{subfigure}
    \begin{subfigure}[b]{0.32\linewidth}
        \centering
        \includegraphics[keepaspectratio=true,width=\linewidth]{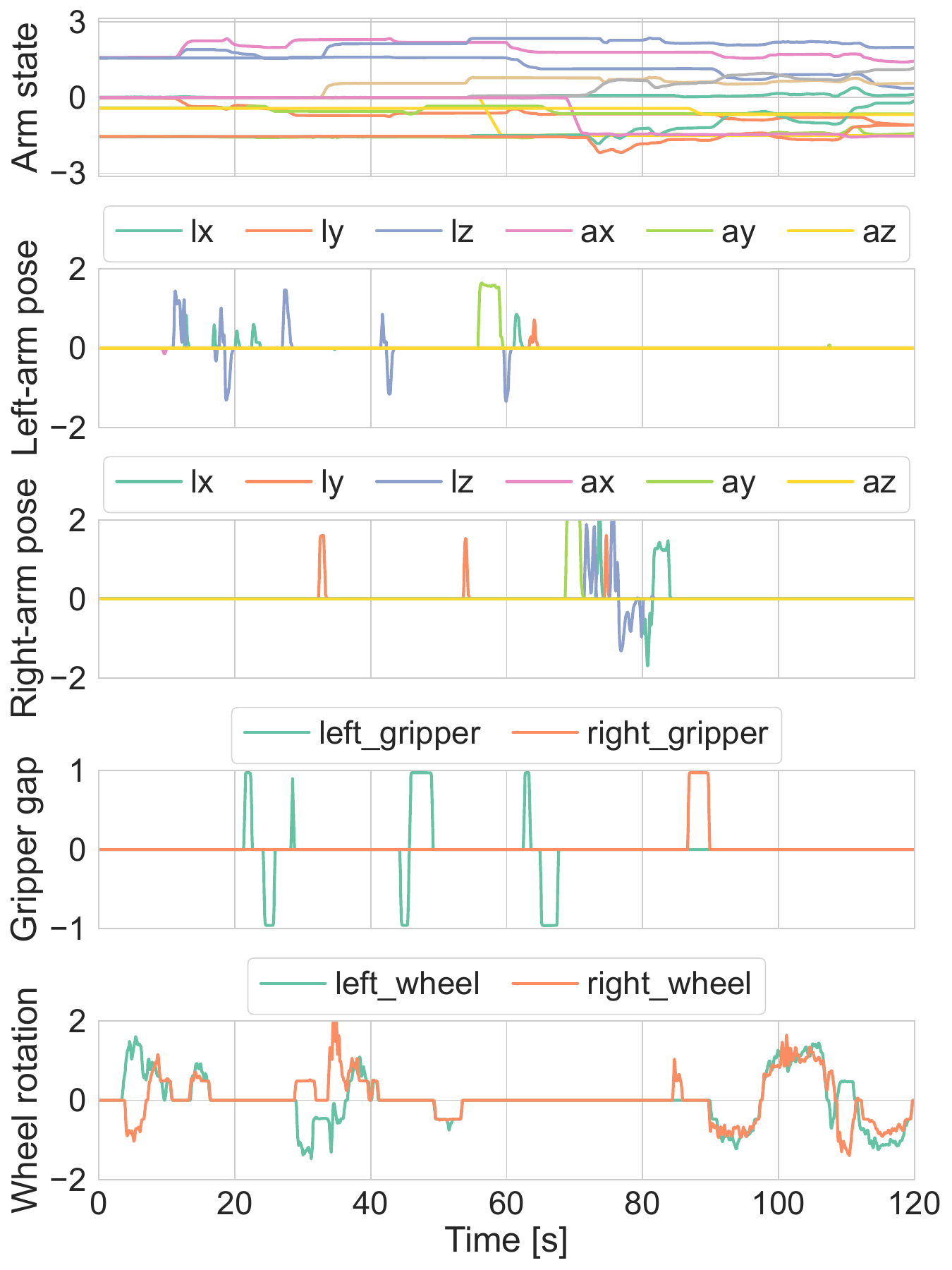}
        \subcaption{\textit{Push cart}}
        \label{fig:demo_push_cart}
    \end{subfigure}
    \caption{Trajectories of demonstrations}
    \label{fig:demo}
\end{figure*}

Now, we can perform various tasks by operating the dual-arm mobile manipulator with the proposed framework.
The following three types of tasks are demonstrated in this paper, as shown in Fig.~\ref{fig:task}.
\begin{itemize}
    \item \textit{Make coffee} as precise manipulation: the robot picks and places a cup in the center of the spout on a coffee machine and presses its small button.
    \item \textit{Move box} as dual-arm manipulation: the robot moves a box, which is held by both arms, by synchronizing the movements of both arms.
    \item \textit{Push cart} as mobile manipulation: the robot picks and places an object on a cart and pushes it with both arms by grasping its handle.
\end{itemize}
In these demonstrations, the model's classification results for the five steps of observed data are averaged with the same weights for smoother operation.
Therefore, control commands are sent to the robot at approximately 20~Hz, although in reality this is slightly less due to the computational costs and communication delays.

The control commands to the robot during the demonstration are summarized in Fig.~\ref{fig:demo}.
From the top, the arm states (i.e., without wheels), the left-arm pose command, right-arm pose command, gripper gap commands, and wheel rotation commands are depicted.
These behaviors are shown in the attached video.
It can be seen that the robot motions were generated smoothly, although there were some misclassifications and/or operational errors.
In \textit{Make coffee}, the operator easily fine-tuned the position, making full use of the redundancy of the mobile base.
In \textit{Move box}, the movements of both arms were successfully synchronized.
In \textit{Push cart}, the model was robust to the distribution shift that should occur with large movements of the mobile base.
Note that the arms were extended in the attached video when the cart was pulled backward, due to the power limit and backdrivability of the robot system used.

\subsection{Discussions}

Although the proposed framework successfully classified the contact actions with high accuracy and used them to operate the robot to demonstrate multiple tasks, several open issues emerged empirically.
The first is the quality of the registered contact motions.
Although this was taken into consideration when defining them in Section~\ref{subsec:data}, the actual operation of the robot required a deal of familiarity, and even an experienced operator occasionally made mistakes.
Even though all robot operations can be completed on the skin sensors, differences in the way of contact tend to result in discontinuous operations.
The intuitiveness is sacrificed in exchange for the high degree of freedom to some extent inevitably, but there should be room to design more sophisticated contact motions and mapping with robot motions from the viewpoints of ergonomics and psychology~\cite{gholami2022quantitative}.
Alternatively, the interface may be used in conjunction with the direct operation mentioned at the introduction, instead of handling all the degrees of freedom with this interface, to bring out the best of both worlds.

The second is the limitation of the classifier.
As a first step toward the new concept of SMI, this paper used a simple classifier, i.e., the input data is assumed to fit into one of the registered contact motion classes.
However, this makes it difficult to synthesize multiple contact motions to generate multi-dimensional robot motions, although the data collection smoothed out the transitions between different contact motions, as explained in Section~\ref{subsec:data}.
In addition, the magnitude of the robot motion could not be determined by the classifier alone, which led to the addition of the heuristic in eq.~\eqref{eq:gain}.
To avoid these problems, a model that embeds the contact motions into the latent space~\cite{kobayashi2022latent}, which can be directly mapped to the classes, might be effective.

Finally, there is the vulnerability to distribution shifts.
Different operators, different patch locations attached, and other factors could result in different time-series data with misclassification, even if the same contact motions were intended.
As described in Section~\ref{subsec:data}, iterative data collection was used to fill in these differences, but a more fundamental solution is required.
However, a larger model and/or dataset is not appropriate for a robot interface, which must be lightweight and fast.
An adaptive framework that directly captures and corrects distributional shifts in the manner of transfer learning~\cite{kumar2024transfer} would be more promising.

Despite these issues, the proposed framework has a high degree of freedom and is highly extensible.
For example, the robot motions that can be associated with the registered contact motions are not limited to primitive ones, as shown in this paper, but a complex sequence of motions (e.g., grasping a nearby object) is allowed.
Such shortcuts would contribute to the semi-automation of robot operations.
Alternatively, although the two patches were treated separately in this paper, it is possible to integrate sensing information from both to determine the operation strategies.
If the robot is working with both arms as in the demonstration, it would be easier to operate the robot by determining whether they should be moved synchronously or not.

\section{Conclusion}

This study proposed a novel Skin Machine Interface (SMI) framework that employs multimodal skin sensors as a robot interface.
The spatiotemporal multimodal measurements facilitate the training of a model capable of classifying diverse human contact motions.
Consequently, this enables the operation of robots with high degrees of freedom using a single interface.
Experiments have demonstrated that multimodality and the soft (and elastic) supports of the sensors, which can fully exploit multimodality, are crucial for enhancing classification accuracy.
The proposed framework was sufficient to perform a variety of tasks with the dual-arm mobile manipulator.
In the future, we will optimize the design of robot/contact motions and strategies to further harness the potential of this interface.

%
%
%
\bibliographystyle{IEEEtran}
{
\bibliography{biblio}
}


\end{document}